\documentclass{article}

\usepackage[final,nonatbib]{neurips_2023}

\usepackage[utf8]{inputenc} % allow utf-8 input
\usepackage[T1]{fontenc}    % use 8-bit T1 fonts
\usepackage{hyperref}       % hyperlinks
\usepackage{url}            % simple URL typesetting
\usepackage{booktabs}       % professional-quality tables
\usepackage{amsfonts}       % blackboard math symbols
\usepackage{nicefrac}       % compact symbols for 1/2, etc.
\usepackage{microtype}      % microtypography
\usepackage{xcolor}         % colors

\usepackage{cite}
\usepackage[math]{alp} 
\usepackage{hyperref}
\usepackage{url}
\usepackage{algorithm}
\usepackage{algorithmic}
\usepackage{multirow}
\usepackage{caption} 
\usepackage{graphicx}
\usepackage{wrapfig,lipsum,booktabs,graphicx}
\usepackage{rotating}
\usepackage{tabularx}
\usepackage{booktabs}

\usepackage{amsmath,amssymb,amsfonts}

\usepackage{graphicx}
\usepackage{textcomp}
\usepackage{xcolor}
\usepackage{subfig}

\title{Enhancing Diffusion-based Point Cloud Generation with Smoothness Constraint}

\author{%
  Yukun Li \\%\thanks{Corresponding email: liping} 
  Department of Computer Science\\
 Tufts University\\
  Medford, MA 02155 \\
  \texttt{yukun.li@tufts.edu} \\
  \And
  Liping Liu \\
  Department of Computer Science\\
  Tufts University \\
  Medford, MA 02155 \\
 \texttt{liping.liu@tufts.edu}\\
}

\begin{document}

\maketitle

\begin{abstract}
Diffusion models have been popular for point cloud generation tasks. Existing works utilize the forward diffusion process to convert the original point distribution into a noise distribution and then learn the reverse diffusion process to recover the point distribution from the noise distribution. However, the reverse diffusion process can produce samples with non-smooth points on the surface because of the ignorance of the point cloud geometric properties. We propose alleviating the problem by incorporating the local smoothness constraint into the diffusion framework for point cloud generation. Experiments demonstrate the proposed model can generate realistic shapes and smoother point clouds, outperforming multiple state-of-the-art methods. 
\end{abstract}

\section{Introduction}
As a widely-used 3D representation, point clouds have recently attracted widespread attention due to their compactness and proximity to raw sensory data \cite{yue2018lidar, mao2019interpolated, hu2020randla, guo2020deep}. Several sophisticated algorithms have been proposed for analyzing point clouds in robotics and autonomous driving \cite{pomerleau2015review, he2023hierarchical, xie2020pointcontrast, he2020curvanet,zhang2022pointclip}. However, acquiring raw data in real applications demands significant labor effort and incurs time costs. Moreover, the obtained point clouds are frequently imperfect; they may be sparse and partial due to factors such as distances, occlusions, reflections, and the limitations of device resolution and angles \cite{zhao2019pointweb,hu2022sqn}. Recently, diffusion-based generative models have been effective in generating 2D images and pose a promising direction for generating 3D point clouds. However, point cloud generation poses more significant challenges than image generation since points in a point cloud have no order and are arranged irregularly.  In addition, finding a way to ensure smoothness on the surface of the point cloud is also a challenging problem.

Several algorithms have been proposed for point cloud generation. For instance, probabilistic methods \cite{yang2019pointflow, luo2021diffusion} take the generation of point clouds as a probabilistic inference problem. Building upon the structure of Variational Autoencoders (VAE), these methods aim to learn the distribution of shapes. The process involves initially learning the distribution of a shape and subsequently learning the conditional distribution of points generated based on this shape. But the Pointflow \cite{yang2019pointflow} needs to normalize the learned distribution, which is time-consuming. ShapeGF \cite{cai2020learning} must require two stages of training. Some methods are based on the autoregressive model \cite{sun2020pointgrow}, but these methods need to specify an order of points during training. \cite{li2018point} proposed a point cloud
GAN model with a hierarchical sampling to generate new
point cloud objects. Other GAN-based methods like \cite{shu20193d, valsesia2018learning, achlioptas2018learning} are challenging to train and unstable, primarily due to the use of the adversarial loss function during the training process.

There are also several diffusion-based generative models \cite{luo2021diffusion, zhou20213d}, which treat points as particles in a thermodynamic system that diffuse from the original distribution to the noise distribution. The generation process can move the noisy point locations toward the original point cloud surface. The training aims to learn a score model to approximate the gradient of log-likelihood of the point cloud distribution, which is equivalent to learning the moving direction of the particles (points) during diffusion. However, both algorithms are discrete generative processes and thus cannot consider the shape of geometric information when learning point-moving directions. Compared with diffusion-based models for image data that learn the data distribution in the feature space, these models face a critical challenge in learning the 3D point location distribution. This paper aims to incorporate the local geometric constraints into a diffusion-based point cloud generation process to enforce surface smoothness. Our approach views the diffusion probabilistic model from a Bayesian perspective to consider the prior constraint of the point cloud. 
Experiments demonstrate that our method is not only capable of generating realistic shapes but also generating smoother point clouds, outperforming multiple state-of-the-art methods.

\section{Background}
\subsection{Tweedie's Formula}
Consider the noisy sample distribution $p(\hat{\bX}|\bX)\sim\mathcal{N}(\bmu,\sigma^2 \bI)$, the Tweedie's formula \cite{robbins1992empirical,kim2021noise2score} tells us the optimum denoised sample can be achieved by computing the posterior expectation:
{\footnotesize \begin{align}
    \mathbb{E}[\bX|{ \hat{\bX}}] =  \hat{\bX} + \sigma^2\nabla_{\hat{\bX}} \log p(\hat{\bX})
\end{align}}
Thus, for a diffusion model in which the forward step is modeled as $p(\hat{\bX}|\bX)\sim\mathcal{N}(a_i \bx_0,b_i^2\bI)$, the formula can approximate the denoised sample $\bx_0$:
{\footnotesize \begin{align}
    \mathbb{E}[\bX_0|{ {\bX_i}}] = ( {\bX_i} + b_i^2\nabla_{{\bX_i}} \log p({\bX_i}))/a_i.
\end{align}}
Here $a_i$ and $b_i$ are the drift and diffusion coefficients of the forward diffusion process. Our approach uses this equation to approximate the clean point cloud from the noisy samples at each time step during the reverse diffusion process.

\subsection{Point Cloud Smoothness Property}
Point clouds have extensive applications in computer vision and robotics. Nevertheless, raw point clouds are often affected by noise and irregularities resulting from data acquisition techniques or environmental factors \cite{he2023survey}. One pivotal approach to improving the quality of point cloud generative models involves implementing smoothness constraints, primarily utilizing graph Laplacian matrices constructed from k-nearest neighbor (KNN) graphs \cite{zeng20193d}. This method characterizes the point cloud's smoothness by quantifying the average difference between each point and its neighboring points, as expressed by the equation:
{\footnotesize \begin{align}\label{eq:smoothness}
S(\bX) = \text{trace}(\bX^T\mathbf{L}\bX)
\end{align}}

Here $\bX$ represents the matrix of all points' positions and $\mathbf{L}$ stands for the Laplacian matrix derived from the KNN graph. Point cloud generative models can attain smoother surfaces by incorporating this smoothness constraint term into an optimization framework.

\section{Approach}
A point cloud consists of a set of points in 3D space. Each point cloud can be denoted by $\bX = \{\boldsymbol{x}_i\}_{i=1}^{N} \in \bbR^{N\times 3}$, where $N$ is the number of points, and $\boldsymbol{x}_i$ is the 3D coordinates for  point $i$. The dataset comprises $M$ distinct point clouds, denoted as $\mathcal{D} = \{ {\bf X}_{1}, ...{\bf X}_{M} \}$. The goal is to develop a model to generate a precise point cloud that faithfully captures the datasets' shape geometry and variability. We  formulate point cloud generation as a continuous particle diffusion process.

\begin{wrapfigure}{r}{7cm}
    \centering
    \includegraphics[width=0.40\textwidth]{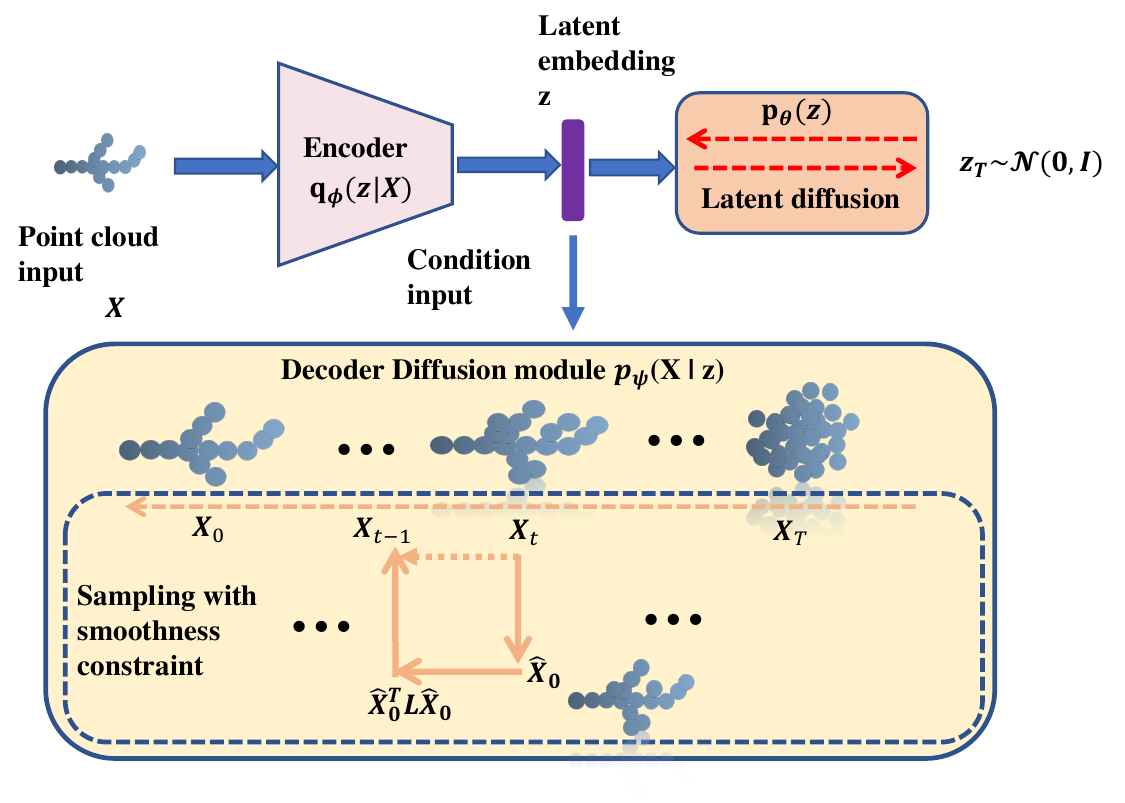}
    \caption{The overall structure of our model, containing three modules: Encoder module $q_{\bphi}(\bz|\bX)$; The latent diffusion learns the latent prior distribution $p_{\btheta}(\bz)$; The decoder diffusion module learns a conditional diffusion model $p_{\boldsymbol{\psi}}{(\bX|\bz)}$ to reconstruct the original shape $\bX$ based on its embedding $\bz$. A smoothness constraint is incorporated during the decoder reverse diffusion process.}
    \label{fig:structure}
\end{wrapfigure}

However, this problem introduces several unique challenges. Firstly, a single object can have different shapes (e.g., various airplane categories).  Therefore, it is essential to model the global shape distribution and the individual point cloud distribution. Secondly, the diffusion probabilistic model captures the intermediate movement of points but may not consider local geometric information within the point cloud, potentially leading to a non-smooth surface during sampling. 
To address these challenges, we propose a diffusion probabilistic model with a smoothness constraint to capture the local geometric features within point clouds effectively. The overall model structure is depicted in
Figure\ref{fig:structure}. The model consists of an encoder, a diffusion decoder, and a latent diffusion module. The encoder $q_{\bphi}(\bf z|\bf X)$ parameterized by $\bphi$ learns a low dimensional latent embedding distribution for the given point cloud input to encode its global shape feature. Given a shape embedding $\bz$, the diffusion decoder $p_{\boldsymbol{\psi}}(\bf X|\bf z)$ reconstructs the original point cloud through a reverse diffusion process. Furthermore, to accurately model the shape latent embedding $\bf z$ prior distribution, we use a latent diffusion module $p_{{\btheta}}(\bf z)$ to transform a standard normal distribution to the aggregated posterior of the encoder.
\subsection{Proposed Overall Model}
Our model learns the point cloud distribution over a diverse shape pattern. As indicated by \cite{li2018point}, for the point cloud generation, it is not helpful to learn the marginal distribution $p(\bX)$ across various shapes. Thereby, we model the distribution into the factorized format and incorporate the latent shape embedding distribution into consideration:  $p_{\btheta,\boldsymbol{\psi}}(\bX, \bz) = p_{\boldsymbol{\psi}}(\bX|\bz)p_{\btheta}(\bz)$,  where  $\bz \in \mathbb{R}^d$ is a latent d-dimensional variable that embeds the global shape pattern of the point cloud,  $p_{\btheta}(\bz)$ represents the latent embedding prior distribution.  $p_{\boldsymbol{\psi}}(\bX|\bz)$ is a conditional diffusion model given a specific global shape condition $\bz$. The model trains the time-dependent score model $s_{\boldsymbol{\psi}}(\bX, \bz, t)$ to learn the score function of $p_{\boldsymbol{\psi}}(\bX|\bz)$, which is the gradient of log-likelihood $\nabla_{\bX}\log p_{\boldsymbol{\psi}}(\bX|\bz) $. The global shape posterior is approximated through another latent diffusion model $q_{\boldsymbol{\phi}}(\bz|\bX)$. We maximize the \textit{Evidence Lower Bound} (ELBO) on the data log-likelihood as Equation \ref{elbo} to optimize the generative model.
% {\scriptsize
\begin{align}
\label{elbo}
\max_{\boldsymbol{\phi}, \boldsymbol{\psi}, \btheta}~\calL(\bphi, \boldsymbol{\psi}, \btheta) &=\mathbb{E}_{q_\bphi(\bz|\bX)}\big[\log{p_{\boldsymbol{\psi}}(\bX|\bz)}\big]  - \KL{q_\bphi(\bz|\bX)}{p_\btheta(\bz)} \nonumber  \\ &=\mathbb{E}_{q_\bphi(\bz|\bX)}\big[\log{p_{\boldsymbol{\psi}}(\bX|\bz)}\big] + \mathbb{E}_{q_\bphi(\bz|\bX)} \big[\log p_\btheta(\bz)\big] +\mathbb{H}\big[q_\bphi(\bz|\bX)\big] 
\end{align}
% }

\subsection{Sampling with Smoothness Constraint}
The framework uses a conditioned diffusion probabilistic model to generate point clouds with diverse shapes. However, it is still challenging to generate the 3D point location distribution obeying the geometrical constraint on the 3D object. 

We propose a graph Laplacian smoothness constraint on the decoder diffusion model  $p_{\boldsymbol{\psi}}({\bf X}|\bz)$. Specifically, we incorporate the point cloud constraint into the sampling process. Suppose a  constraint $\bH$ on the sample takes the form $p(\bH|\bf{X})$, we consider a probabilistic  model from Bayesian perspective:
% {\scriptsize
\begin{align}
     p(\bf{X}|\bH,\bz) = \frac{p(\bH|\bf{X}, \bz) p(\bf{X}|\bz)}{p(\bH,\bz)}   
\end{align}
% }
We propose a Bayesian denoising method by taking the gradient of the  posterior:
% {\scriptsize
\begin{align}
\label{eq:bayes}
\nabla_{\bf{X}} \log p(\bf{X}|\bH,\bz) = \nabla_{\bf{X}} \log p(\bf{X}|\bz) + \nabla_{\bf{X}} \log p(\bH|\bf{X},\bz)
\end{align}
% }

The first term is learned by the decoder diffusion score model $s_{\boldsymbol{\psi}}(\bX, \bz, t)$, and the challenge is how to add the second Bayesian constraints term to each time step of the reverse diffusion process on noisy sample $\bX_t$. To address this challenge, We consider the constraints on the inferred clean point clouds based on Tweedie's formula:
\begin{align} %\scriptsize
  \hat{\bX} = \frac{\bX_{t} + b_t^2\bs_{\boldsymbol{\psi}}(\bX, \bz, t)}{a_t}.
\end{align}
Here $a_t, b_t$ are the drift and diffusion coefficients of the forward diffusion processes. Thus, we can construct a smoothness constraint based on the estimated clean point cloud $\hat{\bf{X}}$. The constraint is proportional to the graph Laplacian regularizer on the KNN graph of the point cloud  to encourage more uniform point distribution during the sampling process:
\begin{align} %\scriptsize
 \log p(\bH|{\bX}, \bz) \propto - \text{trace}( \hat{\bX}^T\mathbf{L}\hat{\bX}), 
 \text{where} \ \hat{\bX} = {\bX_{t} + b_t^2 s_{\boldsymbol{\psi}}(\bX_t, \bz, t)}/{a_t}.
\end{align}

Here $\hat{\bf{X}}$ is the estimated clean point cloud data at each reverse diffusion time step, and $\bf{L}$ is the graph Laplacian matrix.
Putting the Bayesian denoising method together, the reverse diffusion under the  smoothness constraint can be represented by 
% {\scriptsize 
\begin{align}
    &\bX_{{t-1}}=\bX_{{t}} + \underbrace{ [f_t(\bX_t) - g_t^2\bs_{\boldsymbol{\psi}}(\bX_t, \bz, t)]\Delta_t}_{\text{parameterized by score model}}  
    - \underbrace{ \alpha \nabla_{\bX_t} \{ \text{trace} \ (\hat{\bX}_t^T\mathbf{L}\hat{\bX}_t) \} }_{\text{smoothness constraint}}+ g_t(\bf{X}_t)\bz,  \bz\sim\mathcal{N}(0,I). %\nonumber
\end{align}
% }
Here $\alpha$ controls the relative weight of the smoothness constraint, and $f_t, g_t$ are the drift and diffusion coefficients of the forward diffusion process.

\section{Experiment}
This section describes the experimental results of the point cloud generation task. The experiment aims to quantitatively compare our proposed framework with several state-of-the-art  models. 
\subsection{Experiment Setup}
\label{sec:esetup}
{\bf Baseline methods:} We compare our proposed model against several SOTA models for point cloud generation. Those models  include r-GAN\cite{achlioptas2018learning}, l-GAN\cite{valsesia2018learning}, SoftFlow\cite{kim2020softflow}, and DPM\cite{luo2021diffusion}.

{\bf Datasets:} For our experiments, we utilized the  ShapeNet dataset\cite{chang2015shapenet}. We follow \cite{yang2019pointflow} to split dataset.
{\bf Evaluation metrics:}
\label{sec:emetrics}
We adopt the following metrics: Minimum Matching Distance(MMD), Coverage score(COV), 1-NN classifier accuracy (1-NNA), and 
Relative Smoothness (RS). RS is computed by the graph Laplacian as Eq.\ref{eq:smoothness}. It is computed by:
% {\scriptsize 
\begin{align}
    &RS(P_{model}, P_{data}) =\|\mathbb{E}_{X\sim p_{model}} \text{trace}(\bX^T\mathbf{L}\bX) 
    - \mathbb{E}_{\bX\sim p_{data}} \text{trace}(\bX^T\mathbf{L}\bX)\|  
\end{align}
% }
\subsection{Generation Performance Comparison}
\label{sec:eresults}
\begin{wraptable}{r}{7.15cm}\tiny%\scriptsize
\centering
\caption{Comparison of shape generation on ShapeNet. MMD is multiplied by $10^2$. Numbers in bold indicate the best two methods}
\setlength{\tabcolsep}{1mm}
\begin{tabular}{c|clll}
\hline
Category                                      & Model           & MMD ($10^2,\downarrow$) & COV (\%,$\uparrow$) & 1-NNA (\%,$\downarrow$)  \\ \hline
\multicolumn{1}{c|}{\multirow{8}{*}{Airplane}} & r-GAN           & 2.31           & 14.32                          & 96.79                \\
\multicolumn{1}{c|}{}                          & l-GAN            & 0.58           & 21.23                            & 93.95                                      \\
\multicolumn{1}{c|}{}                          & SoftFlow           &0.38            & 47.90                            & \textbf{65.80}                       \\
\multicolumn{1}{c|}{}                          & DPM              & 0.57           & 33.83                         & 86.91                   \\
\multicolumn{1}{c|}{}                          & Ours            & \textbf{0.39}             & \textbf{49.11}                           & 68.89          \\
\multicolumn{1}{c|}{}                          & Ours+constraint & \textbf{0.38}             & \textbf{49.63}                           & \textbf{66.75}                         \\ \hline
\multirow{8}{*}{Chair}                        & r-GAN           & 8.31           & 15.13                            & 99.70            \\
                                              & l-GAN             & 2.01          & 29.31                            & 99.70\\
                                              & SoftFlow            & 1.68           & 47.43                            & {\bf 60.05}\\
                                              & DPM              & 2.07           & 35.50                          & 74.77   \\
                                              & Ours            &{\bf 1.61}                 & {\bf 47.89}                               & 61.15        \\
                                              & Ours+constraint & \textbf{1.60}                 & \textbf{50.19}                                & {\bf 60.54}         \\ \hline
\multirow{8}{*}{Car}                        & r-GAN           &     2.13         &   6.539                          &      99.01        \\
                                              & l-GAN             & 1.23        &    23.58                           & 88.78 \\
                                              & SoftFlow            &   0.86          &     44.60                        & {\bf 60.09}\\                                            
                                              & DPM             &  1.14                &   34.94                              &  79.97     \\
                                              & Ours            & {\bf 0.85}                 &  {\bf 47.16}                               &    71.88      \\
                                              & Ours+constraint & \textbf{0.82}                 & \textbf{47.88}                                &  \textbf{62.90}          \\ \hline
\end{tabular}\label{table:comparison}
\end{wraptable}

We compare our model with baseline models on the airplane, chair, and car categories for point cloud generation. We consider two settings of our proposed model. "Ours" represents the model without smoothness constraint and the other with smoothness constraint. The evaluation results are summarized in Table~\ref{table:comparison}.  
The results show that our proposed model outperforms the GAN and diffusion-based baseline models. Our model performs best in the MMD, COV, and 1-NNA measures. The model with constraint further improves the performance. Thus, it demonstrated the effectiveness of the proposed constraint. We also visualize generated samples in the Appendix.

%\begin{table}[th]\scriptsize
\begin{wraptable}{r}{7.5cm}\scriptsize
\centering
\caption{Comparison of smoothness performance on ShapeNet.  RS is the relative smoothness}
\setlength{\tabcolsep}{1mm}
\begin{tabular}{c|clll}
\hline
Category        & Model           &  RS ($\downarrow$) & GT smoothness  \\ \hline                          \multicolumn{1}{c|}{\multirow{2}{*}{Airplane}}   & Our diffusion model           & 64.90 &\multicolumn{1}{c}{\multirow{2}{*}{1651.92}}\\
\multicolumn{1}{c|}{}                          & Our diffusion model + constraints      &{\bf 37.52}             \\ \hline
\multirow{2}{*}{Chair}   & Our diffusion model                & 1041.35  &\multicolumn{1}{c}{\multirow{2}{*}{4807.79}}  \\
    & Our diffusion model + constraints     &{\bf 983.92} \\ \hline
\multirow{2}{*}{Car}   & Our diffusion model & 203.02  &\multicolumn{1}{c}{\multirow{2}{*}{2677.01}}\\
    & Our diffusion model + constraints    &{\bf 177.90} \\ \hline
\end{tabular}
\label{table:uniform}
\end{wraptable}

% \newpage
\subsection{Smoothness Performance Comparison }\label{sec:unifom}

Furthermore, to demonstrate the performance of the constraint, we use the relative smoothness (RS) metrics to evaluate.  The results are shown in Table~\ref{table:uniform}. The RS metrics measure the relative smoothness between sampled and ground truth point clouds, and we also provide the ground-truth smoothness in the last column. By incorporating the smoothness constraints into the generation model, we can observe a reduction in the relative smoothness metrics. This serves to validate the effectiveness of the proposed smoothness constraint in generating a smoother surface.

\section{Conclusions}
In this paper, we introduce a novel diffusion-based model for point cloud generation, incorporating not only the global geometry distribution but also integrating a local smoothness constraint into the generation process. Experimental results demonstrate that the proposed framework can produce realistic and smoother point cloud samples. The limitation of our approach is that we only consider the smoothness constraint on the point cloud.

\begin{ack}
We thank anonymous reviewers for their valuable feedback. This work is partially supported by the NSF CAREER Award \#2239869.
\end{ack}

% \section{Supplementary Material}

% Authors may wish to optionally include extra information (complete proofs, additional experiments and plots) in the appendix. All such materials should be part of the supplemental material after the references in the main text.

\newpage
\bibliography{ref}
\bibliographystyle{unsrt}
%%%%%%%%%%%%%%%%%%%%%%%%%%%%%%%%%%%%%%%%%%%%%%%%%%%%%%%%%%%%
\newpage
\section*{Appendix}
\section{Background: Diffusion-based Generative Model}
We consider the diffusion process over data $\bX$ indexed with time $t \in [0,1], \{\bX_t\}_{t=0}^1$. Here $\bX_0 \sim p_0(\bX)=p_{data}$, $\bX_1 \sim p_1(\bX)$ , where $p_1(\bX)$ approximates the Gaussian distribution $\mathcal{N}(\bzero, \bI)$. The diffusion process can be modeled by It\^{o} SDE\cite{song2020score}:
\begin{align}\label{eq:sde}
    d\bX_t = f_t(\bX, t)dt + g_td\bw,
\end{align}
$f_t$ is a drift coefficient, and $g_t$ is a diffusion coefficient.
The forward diffusion process smoothly transforms $\bX$ by adding infinitesimal noise $dw$ at each infinitesimal step $dt$.

To generate samples following the data distribution, we start from the prior distribution and follow the reverse diffusion process, which is defined by:
\begin{align}\label{eq:rvsde}
    d\bX_t = [f_t(\bX_t) - g_t^2\nabla_{\bX_t} \log p_t(\bX_t) ]dt + g_td\bar{\bw},
\end{align}
The diffusion model training process learns the time-dependent score model $\bs_\theta(\bX, t)$ and minimizes the least square loss between the score function and the model output
\begin{align}
    \min \int_{0}^{1} \mathbb{E}_{p_t(\bX)}[\|\bs_\theta(\bX, t) - \nabla_{\bX}\log p_t(\bX)\|^2] dt
\end{align}
Thus, the sampling process replace the unknown score function $\nabla_{\bX_t} \log p_t(\bX_t)$ in Eq~\ref{eq:rvsde} with the score model $\bs_\theta(\bX, t)$.
\section{Details on Approach}
% \subsection{Model Architecture}
\subsection{Diffusion-based Prior Model}
The prior model aims to capture the distribution of the global geometry of all point cloud shapes. A common choice is to use a fixed prior (e.g., $\bz \sim \mathcal{N}(\bzero, \bI)$) in variational auto-encoder, but it can produce holes, then there exists a mismatch between the aggregated posterior $\sum_i q_{\boldsymbol{\phi}}(\bz|\bX^{(i)})$ and the prior $p(\bz)$. Then, this can lead to some regions in the latent space that cannot be decoded into data-like samples. 

Our proposed model uses a learnable prior model $p_{\theta}(\bz)$. The prior is flexible, such that it can catch up with the aggregated posterior, and then they can match each other better, thus alleviating the problem of prior holes. Here, we propose to use the diffusion process to learn the prior over the shape distribution.
Specifically, given a global geometry encoding $\bz_0 \sim q_{\boldsymbol{\phi}}(\bz_0|\bX)$, we diffused $z_0$ through a SDE defined in Eq.~\ref{eq:sde} to the standard Normal distribution $p(\bz_1) = \mathcal{N}(\bzero, \bI)$, and we train a time-dependent score model $\bs_{\btheta}(\bz_t, t)$ to approximate the true score $\nabla_{\bz_t}\log p_{\btheta}(\bz_t)$. For the generation process, the reversed SDE based on Eq.~\ref{eq:rvsde} is employed to generate samples that follow the prior distribution $p_{\btheta}(\bz_0)$. We use several layers of 1D convolution with residual connection for the score model. The details are in the experiment section. 

\subsection{Diffusion-based Conditional Decoder}\label{sec:condscore}
In this section, we introduce the framework of the conditional diffusion model in the decoder part. Here, we consider the point cloud generation as a conditional diffusion process, where the condition is the global geometry encoding $\bz \sim q_{\boldsymbol{\phi}}(\bz|\bX)$. Similarly, we diffuse the points of the point cloud (particles) to the standard Normal distribution and learn a conditional score model $\bs_{\boldsymbol{\psi}}(\bX_t, \bz, t)$ to approximate the reverse moving direction of the points during the forward diffusion process (equivalent to the score $\nabla_{\bX_t}\log p_{\boldsymbol{\psi}}(\bX_t|\bz)$). For sampling, given a  global geometry condition $\bz$, we can use the score model $s_{\boldsymbol{\psi}}(\bX_t, \bz, t)$ through the reverse SDE to move the noisy point cloud to the original point cloud distribution.  

Specifically, the input of this network is the perturbed point cloud with Gaussian noise sampled from $\bX_t \sim p_t(\bX_t|\bX)$
and the time embedding. We also concatenate the latent code $\bz$ as the condition to the input of each layer.

\subsection{Training algorithm}
In the training objective of Eq.\ref{elbo}, we estimate the entropy term $\mathbb{H}\big[q_{\boldsymbol{\phi}}(\bz|\bX)\big]$ via Monte Carlo samples. 
And our decoder is a continuous diffusion-based conditional decoder, then the reconstruction term $\mathbb{E}_{q_{\boldsymbol{\phi}}}(\bz|\bX)\big[\log{p_{\boldsymbol{\psi}}(\bX|\bz)}\big]$ is approximated by mean square loss of the score function
\begin{align}
   \mathbb{E}_{q_{\boldsymbol{\phi}}}(\bz|\bX)\big[\log{p_{\boldsymbol{\psi}}(\bX|\bz)}\big] &= \mathbb{E}_{t,p_t(\bX)} \big[\frac{g(t)^2}{2}\|\bs_{\boldsymbol{\psi}}(\bX_t, \bz, t)- \nabla_{\bX}\log p_t(\bX_t|\bz)\|^2_2\big] \big]
\end{align}
and we optimize the cross-entropy term through score matching following \cite{vahdat2021score, chen2022nvdiff} with $t\sim U[0,1]$: 
\begin{align}
  \mathbb{E}_{q_{\boldsymbol{\phi}}(\bz|\bX)}& \big[\log p_\theta(\bz)\big] =\bbE_{t,q_{\bphi}(\bz_0,\bz_t|\bX)}\Big[\frac{g(t)^2}{2}\|\bs_\btheta(\bz_t,t)  - \nabla_{\bz}\log p_t(\bz_t)
  \|^2_2\Big]
\end{align}
The training process is described in Algorithm~\ref{alg:train}
\begin{algorithm}[t]
  \caption{{Training Procedure}}
  \begin{algorithmic}
  \label{alg:train}
    \STATE {\bfseries Input: point cloud training data  
  $p_{data}(\bX)$}
    \FOR{epoch = $0$ to $n$}
        \STATE Sample $  \bX_{0} \sim p_{data}(X)$, $t \sim \text{Uniform}(0,1)$
        \STATE Encoder: $\bz_0 = q_{\bphi}(\bz|\bX_0), \bz_{t} \sim p_t(\bz_t|\bz_0)$
        \STATE Latent diffusion loss: $ L_{\bz}^t \leftarrow  \frac{g(t)^2}{2}\|\bs_\btheta(\bz_t,t) - \nabla_{\bz}\log p_t(\bz_t)
  \|^2_2$   

        \STATE Sample $  \bX_t \sim p_t(\bX_t|\bX_0)$ 
        \STATE Recon loss: $ L_{\bx}^t \leftarrow \frac{g(t)^2}{2}\|\bs_{\boldsymbol{\psi}}(\bX_t, \bz, t) - \nabla_{\bX}\log p_t(\bX_t|\bz)\|^2_2$ 
        \STATE Apply stochastic gradient on the neural network parameters with $ \nabla_{\btheta}(L_{\bz}^t + L_{\bx}^t + \mathbb{H}[q_\bphi(\bz|\bX_t)])$ 
    \ENDFOR
  \end{algorithmic}
\end{algorithm}

\begin{algorithm}[t] 
  \caption{{Sampling Procedure}}
  \begin{algorithmic}
  \label{alg:sampling}
    \STATE {\bfseries Input:} Latent score model $\bs_\btheta(\bz_t, t)$ and decoder score model $\bs_\bphi(\bX_t, \bz, t)$
    \STATE Sample $\bepsilon,\bx_{\text{init}} \sim \mathcal{N}(\bepsilon;\bzero, \bI)$
    \STATE Latent sampling: $\bepsilon \xrightarrow{\text{diffusion}} \bz$
    
    \STATE $  \bX_{t_{n}} = \bX_{\text{init}}$, $t = n$, $\Delta_t = \frac{1}{n}$
    \FOR{$i = n$ to $1$}
      \STATE $ \small \bX_{{t-1}} =\bX_{{t}} + { [f_t(\bX_t) - g_t^2\bs_{\boldsymbol{\psi}}(\bX_t, \bz, t)]\Delta_t} - { \alpha \nabla_{\bX_t} \{ \text{trace} \ (\hat{\bX}_t^T\mathbf{L}\hat{\bX}_t) \} }+ g_t(\bf{X}_t)\bz, \ \bz\sim\mathcal{N}(\bzero,\bI)$
    \ENDFOR
    \RETURN $\bX_{0}$
  \end{algorithmic}
\end{algorithm}

 \section{More Details on Experiments}

 \subsection{Experiment Setup}
 {\bf Datasets:} In our experimental evaluation, we employed the widely used ShapeNet dataset\cite{chang2015shapenet}, a comprehensive benchmark for 3D shape understanding. The dataset comprises 51,127 shapes distributed across 55 different categories. To maintain consistency with prior work and ensure fair comparisons, we adopted the same dataset split as presented in \cite{yang2019pointflow}. 

Within the ShapeNet dataset, e each 3D shape is depicted as a point cloud consisting of 2048 points uniformly sampled on the object's surfaces. This representation of point clouds effectively captures the intricate geometric details of the shapes, making it a fitting input for our 3D shape understanding tasks. 

For our comparative analysis, we specifically focused on three categories: airplanes, chairs, and cars. These categories represent different types of objects with varying shapes and complexities, making them suitable for assessing the performance of our model under diverse scenarios.
\subsection{Implementation Detail}
{\bf Hyperparameters:} The latent code dimension is set to 256. For the training stage, we set the batch size as 32. We utilized Adam optimizer with learning rate $2\times10^{-3}$ for the encoder and learning rate $2\times10^{-4}$ for the diffusion decoder module, $1\times 10^{-4}$ for the latent diffusion module. The learning rate was kept constant for the first 1000 epochs and decreased linearly from 1000 to 2000 epochs. For sampling, We sample each point cloud with N = 2048 points. The time step is set to be $1000$ steps for sampling. 

{\bf Model architecture:} We use PointNet~\cite{qi2017pointnet} as an encoder to map the input point cloud into a 512-dimension latent feature code $\bz$. For the decoder and the prior diffusion module, we utilize OccNet~\cite{mescheder2019occupancy}. We stacked 6 ResNet blocks with 256 dimensions for every hidden layer.

{\bf Evaluation metrics:}

\label{sec:emetrics}
Suppose $P_r$ is the set of reference point clouds for the generation task, and $P_g$ is the set of generated point clouds. We adopt the following metrics:

\begin{enumerate}
    \item Minimum Matching Distance(MMD)
\begin{equation*}\footnotesize
    MMD(P_r, P_g) = \frac{1}{|P_r|}\sum_{P \in P_r}\min_{Q \in P_g} D(P, Q)
\end{equation*}
MMD calculates the average of the closest distances to each reference point cloud to measure the quality of the generated samples. 
\item Coverage score(COV)  
\begin{align*}\footnotesize 
    COV(P_r, P_g) = \frac{|\{\arg min_{P \in P_r} D(P,Q)|Q\in P_g\}|}{|P_r|}
\end{align*}
COV measures the variety of the generated samples to detect mode collapse.
\item
1-NN classifier accuracy (1-NNA): The score is computed using a 1-NN classifier to classify the generated samples and the reference point cloud. When the quality of the generated samples is high, the 1-NNA score should be close to 50\%.
\item 
Relative Smoothness (RS): The smoothness of the sampled point cloud is computed by the graph Laplacian as Eq.\ref{eq:smoothness}. It is computed by:
\begin{align*}\footnotesize 
    &RS(P_{model}, P_{data}) = \|\mathbb{E}_{X\sim p_{model}} \text{trace}(\bX^T\mathbf{L}\bX) 
    - \mathbb{E}_{\bX\sim p_{data}} \text{trace}(\bX^T\mathbf{L}\bX)\|  
\end{align*}

\end{enumerate}
\begin{figure}
    \centering
    \includegraphics[width=0.28\textwidth]{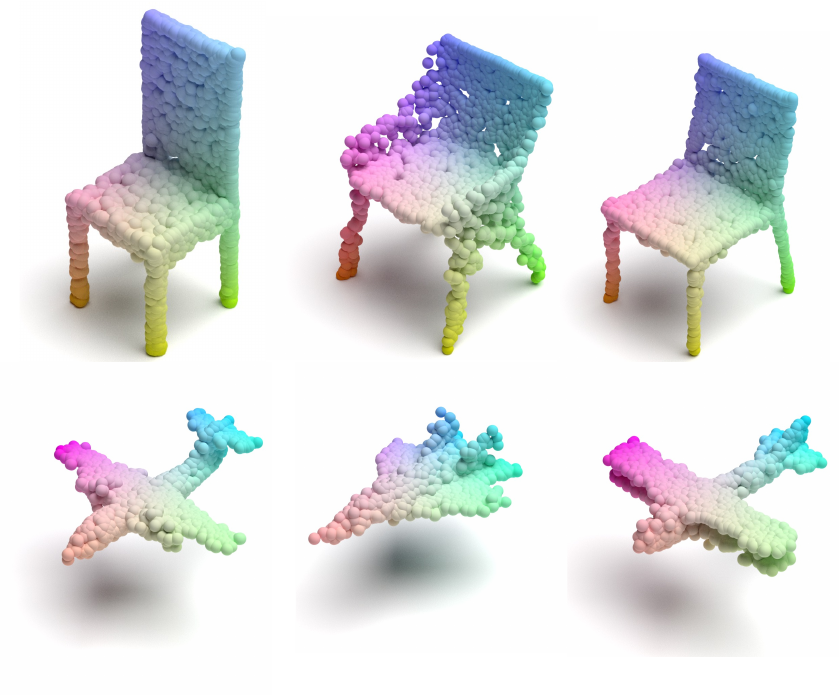}
    \caption{Visualization of some generated samples by our model}
    \label{fig:samples}
\end{figure}

\begin{wrapfigure}{r}{6cm}
    \centering
\includegraphics[width=0.37\textwidth]{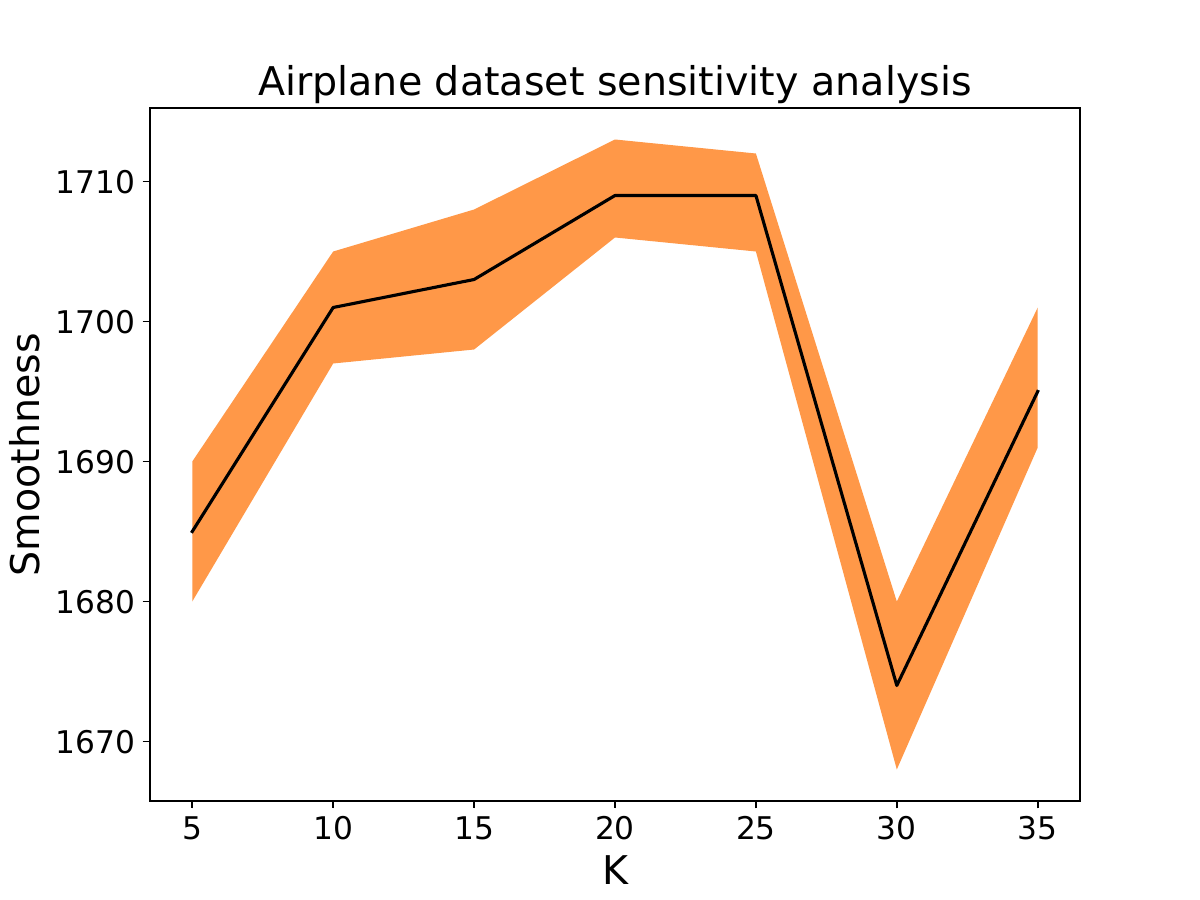}
    \caption{Sensitivity analysis on varying $K$ in KNN graph construction for airplane dataset}
    \label{fig:sensitity}
\end{wrapfigure}

In addition, we perform a sensitivity analysis to investigate how the choice of neighbors of the KNN graph influences the smoothness performance. Specifically, we conduct experiments by varying $K$ from $5$ to $35$ to construct the KNN graph on the airplane datasets. We observe that the sample smoothness is constantly lower than that generated by the diffusion model without smoothness constraint ($1715$), and the best result is achieved when $K=30$.

\end{document}